\documentclass{article}
\usepackage{float}
\usepackage{graphicx}
\usepackage{enumitem}
\usepackage{mathtools}
\usepackage{amsmath}
\usepackage{dsfont}
\usepackage{amssymb}
\usepackage{xcolor}
\usepackage{multirow}
\usepackage{booktabs}
\usepackage{hyperref}
\usepackage{amsmath,amsfonts,amssymb}

\usepackage[final]{neurips_2021}

\DeclareMathOperator*{\argmax}{arg\,max}

\title{Nearest-neighbour baselines for fingerprint prediction from MS/MS spectra under different assumptions}
\author{Ling Min Serena Khoo \\
        \texttt{klingmin@mit.edu}}
\begin{document}

\maketitle

\section{Introduction}
It has recently been shown that nearest-neighbour retrieval provides a strong baseline for molecular fingerprint prediction from MS/MS spectra, with several variants matching or outperforming current deep learning models~\cite{khoo2026why, liu2026massspecgymwilduncoveringcorrecting, Gupta2026-oq}. Importantly, ``nearest neighbour'' encompasses a family of retrieval methods that differ in the information assumed to be available at inference time. In this report, we systematically compare several nearest-neighbour variants, making explicit the information available to each method at inference time and demonstrating how these differing assumptions affect their performance. We consider three assumptions, each giving the retrieval strictly more information than the previous one:

\begin{enumerate}[leftmargin=*]
    \item \textbf{No chemical formula.} Only the spectrum (and its precursor $m/z$) is known. The test spectrum is compared with every training spectrum by a spectral similarity, and the fingerprint of the best match is returned.
    \item \textbf{Chemical formula.} The chemical formula of the test compound is known. Retrieval is restricted to the training spectra with the same formula; when there are none, the search falls back to the whole training set as under assumption~1.
    \item \textbf{Peaks annotated with the ground-truth chemical formula.} The chemical formula is known and is used to annotate each measured peak with a sub-formula of the precursor ion. The formula is the ground truth, but the peak annotations themselves are predicted (by mass decomposition) rather than known. Formula-annotated fragments and neutral losses can be compared between spectra of different formulas, so test spectra without a same-formula training spectrum are retrieved by this sub-formula rule instead. We also apply the sub-formula rule to every test spectrum.
\end{enumerate}

Assumption~3 corresponds to the type of input used by current state-of-the-art fingerprint prediction models. MIST~\cite{goldman2023mist}, for example, takes as input spectral peaks annotated with sub-formulas derived from the known precursor formula. In our implementation of NN, we selected a total of 100 peaks. The number of peaks retained is a modeling choice that may vary across methods; nevertheless, we emphasize that the type of information available under this setting is consistent with that used by current fingerprint prediction models.

Under each assumption we compare five spectral similarities (cosine, modified cosine and neutral-loss cosine as implemented in matchms~\cite{dejonge2024matchms}, the cosine between DreaMS embeddings~\cite{bushuiev2025dreams}, and the cosine between binned spectra) on two datasets, NPLIB1~\cite{duhrkop2021canopus} and MassSpecGym~\cite{bushuiev2024massspecgym}, each under a scaffold and a random split. All methods are described in detail in the Supplementary Information.

\section{Results}

Table~\ref{tab:jaccard} compares the methods by the binary Jaccard index (Tanimoto similarity) between the predicted and the true fingerprint, and Table~\ref{tab:cosine} by the cosine similarity between the raw prediction and the true fingerprint, which does not require binarising the prediction (Supplementary Note~S1). Every nearest-neighbour method returns a prediction for every test spectrum.

\begin{table}[]
\centering
\small
\begin{tabular}{llcccc}
\toprule
& & \multicolumn{2}{c}{\textbf{NPLIB1}} & \multicolumn{2}{c}{\textbf{MassSpecGym}} \\
\cmidrule(lr){3-4} \cmidrule(lr){5-6}
\textbf{Assumption} & \textbf{Similarity} & \textbf{Scaffold} & \textbf{Random} & \textbf{Scaffold} & \textbf{Random} \\
\midrule
\multirow{5}{*}{\shortstack[l]{(1) No chemical\\formula}} & Cosine & 0.208 & 0.519 & 0.235 & 0.740 \\
 & Modified cosine & 0.236 & 0.517 & 0.246 & 0.700 \\
 & Neutral-loss cosine & 0.187 & 0.533 & 0.220 & 0.749 \\
 & Cosine Similarity (DreaMS) & 0.263 & 0.691 & 0.299 & 0.895 \\
 & Binned cosine & 0.178 & 0.592 & 0.231 & 0.850 \\
\midrule
\multirow{5}{*}{\shortstack[l]{(2) Chemical\\formula}} & Cosine & 0.221 & 0.717 & 0.278 & 0.933 \\
 & Modified cosine & 0.242 & 0.732 & 0.291 & 0.934 \\
 & Neutral-loss cosine & 0.206 & 0.707 & 0.268 & 0.936 \\
 & Cosine Similarity (DreaMS) & 0.259 & 0.745 & 0.308 & 0.945 \\
 & Binned cosine & 0.197 & 0.696 & 0.258 & 0.929 \\
\midrule
\multirow{6}{*}{\shortstack[l]{(3) Formula-\\annotated peaks}} & Cosine + sub-formula & 0.309 & 0.753 & 0.365 & 0.937 \\
 & Modified cosine + sub-formula & 0.309 & 0.752 & 0.365 & 0.937 \\
 & Neutral-loss cosine + sub-formula & 0.308 & 0.751 & 0.364 & 0.942 \\
 & Cosine Similarity (DreaMS) + sub-formula & 0.311 & \textbf{0.758} & 0.371 & \textbf{0.947} \\
 & Binned cosine + sub-formula & 0.312 & 0.752 & 0.365 & 0.935 \\
 & Sub-formula rule only & \textbf{0.337} & 0.757 & \textbf{0.381} & 0.937 \\
\bottomrule
\end{tabular}
\vspace{2mm}
\caption{Binary Jaccard index (Tanimoto similarity) between the predicted and the true fingerprint (Morgan, radius 2, 4096 bits). Predictions are binarised at 0.5. Mean over the full test split. \emph{+ sub-formula}: same-formula retrieval with the given similarity, sub-formula rule for test spectra without a same-formula training spectrum. Best value per column in bold.}
\label{tab:jaccard}
\end{table}

\begin{table}[]
\centering
\small
\begin{tabular}{llcccc}
\toprule
& & \multicolumn{2}{c}{\textbf{NPLIB1}} & \multicolumn{2}{c}{\textbf{MassSpecGym}} \\
\cmidrule(lr){3-4} \cmidrule(lr){5-6}
\textbf{Assumption} & \textbf{Similarity} & \textbf{Scaffold} & \textbf{Random} & \textbf{Scaffold} & \textbf{Random} \\
\midrule
\multirow{5}{*}{\shortstack[l]{(1) No chemical\\formula}} & Cosine & 0.313 & 0.593 & 0.331 & 0.777 \\
 & Modified cosine & 0.346 & 0.598 & 0.347 & 0.745 \\
 & Neutral-loss cosine & 0.290 & 0.602 & 0.311 & 0.781 \\
 & Cosine Similarity (DreaMS) & 0.380 & 0.754 & 0.400 & 0.913 \\
 & Binned cosine & 0.276 & 0.650 & 0.315 & 0.869 \\
\midrule
\multirow{5}{*}{\shortstack[l]{(2) Chemical\\formula}} & Cosine & 0.331 & 0.768 & 0.372 & 0.945 \\
 & Modified cosine & 0.356 & 0.784 & 0.388 & 0.946 \\
 & Neutral-loss cosine & 0.315 & 0.759 & 0.359 & 0.947 \\
 & Cosine Similarity (DreaMS) & 0.377 & 0.797 & 0.407 & 0.955 \\
 & Binned cosine & 0.302 & 0.746 & 0.346 & 0.941 \\
\midrule
\multirow{6}{*}{\shortstack[l]{(3) Formula-\\annotated peaks}} & Cosine + sub-formula & 0.476 & 0.820 & 0.519 & 0.952 \\
 & Modified cosine + sub-formula & 0.476 & 0.818 & 0.519 & 0.952 \\
 & Neutral-loss cosine + sub-formula & 0.475 & 0.817 & 0.518 & 0.955 \\
 & Cosine Similarity (DreaMS) + sub-formula & 0.479 & 0.824 & 0.524 & \textbf{0.960} \\
 & Binned cosine + sub-formula & 0.479 & 0.819 & 0.518 & 0.949 \\
 & Sub-formula rule only & \textbf{0.528} & \textbf{0.842} & \textbf{0.556} & 0.952 \\
\bottomrule
\end{tabular}
\vspace{2mm}
\caption{Cosine similarity between the raw prediction (the retrieved fingerprint, the sub-formula consensus before thresholding) and the true fingerprint. Mean over the full test split. \emph{+ sub-formula}: same-formula retrieval with the given similarity, sub-formula rule for test spectra without a same-formula training spectrum. Best value per column in bold.}
\label{tab:cosine}
\end{table}

\paragraph{No chemical formula.} Without the formula, retrieval with DreaMS embeddings is the strongest method on every dataset and split under both metrics (Jaccard index 0.263, 0.691, 0.299 and 0.895 on NPLIB1 scaffold, NPLIB1 random, MassSpecGym scaffold and MassSpecGym random). The choice of spectral similarity matters under this assumption: the five similarities span up to 0.195 in Jaccard index.

\paragraph{Chemical formula.} Restricting retrieval to training spectra of the same formula helps most where most test formulas occur in the training set, i.e. on the random splits (80.7\% of NPLIB1 and 97.1\% of MassSpecGym test spectra have a same-formula training spectrum). Under the scaffold splits this holds for only 38.2\% and 41.0\% of the test spectra, and the remaining spectra, which fall back to the whole training set, are predicted poorly (Jaccard index 0.135--0.236 on NPLIB1 and 0.119--0.194 on MassSpecGym across similarities).

\paragraph{Formula-annotated peaks.} Using the formula to annotate the peaks gives the best results on every dataset and split (best Jaccard index 0.337, 0.758, 0.381 and 0.947; best cosine similarity 0.528, 0.842, 0.556 and 0.960). The gain over assumption~2 is largest on the scaffold splits (+0.078 and +0.073 in Jaccard index on NPLIB1 and MassSpecGym, against +0.014 and +0.003 on the random splits), where most test spectra lack a same-formula training spectrum and the sub-formula rule replaces the weak whole-set fallback. Applying the sub-formula rule to every test spectrum is the best method on both scaffold splits under both metrics; on the random splits it is close to the variants that retrieve by spectral similarity within the same formula. Under this assumption the choice of spectral similarity hardly matters (the five \emph{+ sub-formula} variants lie within 0.013 in Jaccard index on every split).

\appendix

\section*{Supplementary Information}

\subsection*{S1. Data, splits and evaluation}

\paragraph{Datasets and splits.} We use NPLIB1~\cite{duhrkop2021canopus} and MassSpecGym~\cite{bushuiev2024massspecgym}, each under a scaffold split and a random split into training, validation and test sets (Table~\ref{tab:splits}). All retrieval is against the training split. The validation split is used only to select the four hyperparameters of the sub-formula rule (Note~S5), and the test split is never used for any choice.

\begin{table}[H]
\centering
\small
\begin{tabular}{llrrrccc}
\toprule
\textbf{Dataset} & \textbf{Split} & \textbf{Train} & \textbf{Val} & \textbf{Test} & \textbf{Coverage} & \textbf{Leakage} & \textbf{FP oracle} \\
\midrule
NPLIB1 & Scaffold & 12,909 & 2,688 & 2,689 & 38.2\% & 1.6\% & 0.523 \\
NPLIB1 & Random & 12,800 & 2,742 & 2,744 & 80.7\% & 62.9\% & 0.870 \\
MassSpecGym & Scaffold & 76,243 & 16,041 & 16,042 & 41.0\% & 14.3\% & 0.562 \\
MassSpecGym & Random & 75,828 & 16,248 & 16,250 & 97.1\% & 95.3\% & 0.977 \\
\bottomrule
\end{tabular}
\vspace{2mm}
\caption{Datasets and splits. \emph{Coverage}: test spectra whose neutral formula occurs in the training split (the spectra that same-formula retrieval can serve). \emph{Leakage}: test spectra whose structure (first InChIKey block) occurs in the training split. \emph{FP oracle}: mean over test spectra of the highest Tanimoto similarity between the true fingerprint and any training fingerprint, i.e. the ceiling of any method that returns a single training fingerprint.}
\label{tab:splits}
\end{table}

\paragraph{Fingerprints.} The target of every method is the RDKit Morgan fingerprint of radius 2 with 4096 bits, computed from the SMILES of the compound.

\paragraph{Metrics.} Each method produces a raw prediction $\mathbf{r}(q) \in [0,1]^{4096}$ for a test spectrum $q$ with true fingerprint $\mathbf{y}(q) \in \{0,1\}^{4096}$. For methods that return the fingerprint of a single training compound, $\mathbf{r}$ is that binary fingerprint; for the sub-formula rule it is the weighted average of the top-$k$ training fingerprints (Note~S5). We report
\begin{equation*}
\mathrm{Jaccard}(q) = \frac{\lvert \hat{\mathbf{y}} \wedge \mathbf{y} \rvert}{\lvert \hat{\mathbf{y}} \vee \mathbf{y} \rvert}, \quad \hat{\mathbf{y}} = \mathds{1}[\mathbf{r} \ge 0.5],
\qquad
\mathrm{cosine}(q) = \frac{\mathbf{r}^{\top}\mathbf{y}}{\lVert \mathbf{r} \rVert\, \lVert \mathbf{y} \rVert},
\end{equation*}
averaged over the test spectra. The Jaccard index is computed on the binarised prediction and the cosine similarity on the raw prediction, without a threshold; for a binary prediction the cosine similarity equals $\lvert \hat{\mathbf{y}} \wedge \mathbf{y} \rvert / \sqrt{\lvert \hat{\mathbf{y}} \rvert\,\lvert \mathbf{y} \rvert}$.

\subsection*{S2. Spectral similarities}

Let a spectrum $x$ consist of peaks $\{(m_i, I_i)\}$ with $m/z$ values $m_i$ and normalised intensities $I_i$, and let $p(x)$ denote its precursor $m/z$.

\paragraph{Peak-matching similarities.} The cosine, modified cosine and neutral-loss cosine similarities follow matchms~\cite{dejonge2024matchms} (\texttt{CosineGreedy}, \texttt{ModifiedCosine} and \texttt{NeutralLossesCosine}). For two spectra $x$ and $y$, a set of candidate peak pairs $(i, j)$ is formed, the pairs are sorted by the intensity product $I_i I_j$, and pairs are accepted greedily such that every peak is used at most once. With $\mathcal{A}$ the accepted pairs,
\begin{equation*}
s(x, y) = \frac{\sum_{(i,j) \in \mathcal{A}} I_i I_j}{\sqrt{\sum_i I_i^2}\, \sqrt{\sum_j I_j^2}}.
\end{equation*}
The three similarities differ only in the candidate pairs, with tolerance $\tau = 0.1$\,Da and $\Delta = p(x) - p(y)$:
\begin{itemize}[leftmargin=*]
    \item \emph{Cosine}: $\lvert m_i - m_j \rvert \le \tau$. No filter on the precursor $m/z$ is applied (open search).
    \item \emph{Modified cosine}~\cite{watrous2012networking}: $\lvert m_i - m_j \rvert \le \tau$ or $\lvert m_i - (m_j + \Delta) \rvert \le \tau$, so fragments that are shifted by the difference of the precursor masses also match.
    \item \emph{Neutral-loss cosine}: peaks at or above the precursor $m/z$ are removed, and $\lvert m_i - (m_j + \Delta) \rvert \le \tau$, i.e. the neutral losses $p(x) - m_i$ and $p(y) - m_j$ agree within $\tau$.
\end{itemize}
We use the matchms defaults (tolerance 0.1\,Da, intensities raised to the power 1, $m/z$ to the power 0) on the stored normalised intensities. For the few NPLIB1 spectra with more than 500 peaks (some are close to profile data, with up to 361{,}000 peaks), only the 500 most intense peaks are kept; MassSpecGym spectra have at most 299 peaks and are unaffected. To score every test spectrum against every training spectrum (about $1.2 \times 10^9$ pairs on MassSpecGym), the three similarities were re-implemented as a compiled kernel that follows the matchms code path pair by pair; before every search, 600 pairs are scored with matchms itself and the search is aborted on any difference (the largest difference observed was below $10^{-9}$).

\paragraph{DreaMS.} Each spectrum is embedded with the self-supervised DreaMS model~\cite{bushuiev2025dreams} (pre-trained checkpoint, without contrastive fine-tuning), and spectra are compared by the cosine similarity of their embeddings.

\paragraph{Binned cosine.} Each spectrum is binned at 0.25\,Da up to 2000\,Da by summing the intensities within each bin, with the precursor $m/z$ added as a peak of intensity 100, and spectra are compared by the cosine similarity of the binned vectors.

\subsection*{S3. Assumption 1: no chemical formula}

For a test spectrum $q$ and a similarity $s$, the prediction is the fingerprint of the most similar training spectrum,
\begin{equation*}
\hat{t}(q) = \argmax_{t \in \mathcal{C}(q)} \; s(q, t), \qquad \mathbf{r}(q) = \mathrm{FP}(\hat{t}(q)),
\end{equation*}
with candidate pool $\mathcal{C}(q) = \mathcal{T}$, the whole training split. The formula is not used at any point. This is the nearest-neighbour baseline of~\cite{Gupta2026-oq}, which there ranks the candidate structures of a retrieval task by their similarity to $\mathrm{FP}(\hat{t}(q))$; here the retrieved fingerprint itself is evaluated. Ties are broken by the order of the training split.

\subsection*{S4. Assumption 2: chemical formula}

Let $F(x)$ be the neutral chemical formula of the compound behind spectrum $x$. The candidate pool is restricted to the training spectra with the formula of the test spectrum when there are any,
\begin{equation*}
\mathcal{C}(q) = \begin{cases} \{t \in \mathcal{T} : F(t) = F(q)\} & \text{if this set is non-empty,} \\ \mathcal{T} & \text{otherwise,} \end{cases}
\end{equation*}
and the prediction is the fingerprint of the most similar spectrum in $\mathcal{C}(q)$ under the same similarity as in Note~S3. The first case is the nearest-neighbour baseline of our original Comment~\cite{khoo2026why}; the second makes the method return a prediction for every test spectrum. With the peak-matching similarities, a test spectrum may share no peak with any training spectrum of its formula; all scores are then zero and the first training spectrum of that formula is returned. This affects at most 272 of the 16{,}042 MassSpecGym scaffold test spectra (neutral-loss cosine) and at most 26 on the other splits.

\subsection*{S5. Assumption 3: formula-annotated peaks}

Under assumption~3 the chemical formula is used to annotate the peaks, which yields a similarity that is meaningful between spectra of different formulas. Let $F^{+}(x)$ denote the precursor ion formula and $\mathbf{e}(F)$ the vector of element counts of a formula $F$.

\paragraph{Sub-formula similarity $s_{SF}$.} The 100 most intense peaks of each spectrum are annotated by mass decomposition against $F^{+}$: each peak is assigned the sub-formula $f \le F^{+}$ whose monoisotopic mass is closest to its $m/z$ within $\max(0.01\,\mathrm{Da}, 20\,\mathrm{ppm})$, subject to a loose valence check; peaks without an admissible sub-formula are discarded. The formula is the ground truth, but the annotation of each peak is predicted. Each annotated peak with normalised intensity \(I\) generates a fragment token \(f\) and a complementary neutral-loss token \(F^{+}-f\), each with weight \(\sqrt{I}\); if multiple peaks generate the same token, its weight is the maximum \(\sqrt{I}\) among those peaks. A spectrum is represented by the L2-normalised sparse vector $\mathbf{v}(x)$ over these tokens, and $s_{SF}(q,t) = \mathbf{v}(q)^{\top}\mathbf{v}(t)$. Because the tokens are formulas rather than masses, fragments and losses are comparable between spectra of different precursor formulas.

\paragraph{Formula similarity $s_{F}$.} An element-weighted Bray--Curtis similarity of the element-count vectors of the neutral formulas,
\begin{equation*}
s_{F}(q,t) = 1 - \frac{\sum_{e} w_{e}\, \lvert e_{q} - e_{t} \rvert}{\sum_{e} w_{e}\, (e_{q} + e_{t})} \in [0, 1],
\qquad w_{e} = \begin{cases} w & e \notin \{\mathrm{C}, \mathrm{H}\} \\ 1 & e \in \{\mathrm{C}, \mathrm{H}\}, \end{cases}
\end{equation*}
where $e_q$ and $e_t$ are the counts of element $e$ in $F(q)$ and $F(t)$. It acts as a soft prior towards compositionally close training compounds, and equals 1 for identical formulas. The heteroatom weight $w \ge 1$ makes a mismatch in a heteroatom count $w$ times as costly as a mismatch in carbon or hydrogen, since analogues that share a scaffold typically differ in C and H but keep their heteroatom composition; $w = 1$ recovers the plain Bray--Curtis similarity.

\paragraph{Sub-formula rule.} Training spectra are scored by $\mathrm{score}(q,t) = s_{SF}(q,t) + \alpha\, s_{F}(q,t)$. Instead of copying the fingerprint of the single best match, the raw prediction is a softmax-weighted average of the fingerprints of the $k$ highest-scoring training spectra $t_{(1)}, \dots, t_{(k)}$ with scores $s_{(1)} \ge \dots \ge s_{(k)}$,
\begin{equation*}
\mathbf{r}(q) = \sum_{i=1}^{k} w_i\, \mathrm{FP}(t_{(i)}), \qquad w_i \propto \exp\big((s_{(i)} - s_{(1)}) / T\big), \quad \sum_i w_i = 1,
\end{equation*}
which is binarised at 0.5 for the Jaccard index. With $k = 1$ this reduces to copying the best match; the temperature $T$ controls how sharply the weight concentrates on the top of the ranking.

\paragraph{Two variants.} (i) \emph{+ sub-formula}: test spectra with a same-formula training spectrum are retrieved as under assumption~2 with the given spectral similarity, and the remaining test spectra by the sub-formula rule over the whole training split. The rule uses no spectral similarity, so its predictions for these spectra are identical for the five variants. (ii) \emph{Sub-formula rule only}: every test spectrum is retrieved by the sub-formula rule over the whole training split. There is no hard formula filter; the formula enters only through $\alpha\, s_F$.

\paragraph{Hyperparameter selection.} The four hyperparameters $(w, \alpha, T, k)$ are selected separately for each dataset and split on the validation split, with the protocol used at test time: validation spectra are retrieved against the training split, and the configuration with the highest mean Jaccard index is chosen from the grid $w \in \{1, 2, 4, 8, 16\}$, $\alpha \in \{0, 0.25, 0.5, 1, 2, 4\}$, $T \in \{0.01, 0.02, 0.05, 0.1, 0.2, 0.5\}$, $k \in \{1, 3, 5, 10, 20, 50\}$ (ties broken towards smaller $k$, then larger $T$, smaller $\alpha$ and smaller $w$). For variant (i) the selection uses the validation spectra without a same-formula training spectrum (1{,}409, 547, 8{,}875, 530 spectra for NPLIB1 scaffold, NPLIB1 random, MassSpecGym scaffold and MassSpecGym random), i.e. the spectra the rule is applied to; for variant (ii) it uses all validation spectra. The selected values are listed in Table~\ref{tab:hparams}. The mass tolerances, the 100-peak cut-off and the $\sqrt{I}$ token weighting were fixed a priori, and no parameters are learned.

\begin{table}[H]
\centering
\small
\begin{tabular}{llcccccccc}
\toprule
& & \multicolumn{4}{c}{\textbf{Fallback} (+ sub-formula)} & \multicolumn{4}{c}{\textbf{Sub-formula rule only}} \\
\cmidrule(lr){3-6} \cmidrule(lr){7-10}
\textbf{Dataset} & \textbf{Split} & $w$ & $\alpha$ & $T$ & $k$ & $w$ & $\alpha$ & $T$ & $k$ \\
\midrule
NPLIB1 & Scaffold & 4 & 1 & 0.05 & 10 & 4 & 1 & 0.05 & 10 \\
NPLIB1 & Random & 4 & 1 & 0.1 & 5 & 2 & 4 & 0.05 & 10 \\
MassSpecGym & Scaffold & 8 & 1 & 0.05 & 20 & 8 & 1 & 0.05 & 20 \\
MassSpecGym & Random & 4 & 2 & 0.05 & 10 & 4 & 4 & 0.01 & 5 \\
\bottomrule
\end{tabular}
\vspace{2mm}
\caption{Hyperparameters of the sub-formula rule selected on the validation split: on the validation spectra without a same-formula training spectrum when the rule is the fallback, and on all validation spectra when it is applied to every test spectrum.}
\label{tab:hparams}
\end{table}

\subsection*{S6. Reproducibility}
We make the code for the mentioned nearest-neighbour implementations publicly available \href{https://github.com/serenaklm/ML_MS_analysis}{on GitHub} to facilitate further benchmarking. Instructions for downloading the artifacts used in our evaluations are also provided in the repository. All results are produced by \texttt{benchmarked\_models/nearest\_neighbour/run\_nn.py}, whose settings correspond to the rows of Tables~\ref{tab:jaccard} and~\ref{tab:cosine}: \texttt{<sim>\_no\_formula\_filter} (assumption~1), \texttt{<sim>\_with\_formula\_filter} (assumption~2), \texttt{<sim>\_subformula\_match} (assumption~3, \emph{+ sub-formula}) and \texttt{subformula\_only} (assumption~3, sub-formula rule only), with \texttt{<sim>} one of \texttt{cosine}, \texttt{modified\_cosine}, \texttt{neutral\_loss}, \texttt{dreams} and \texttt{binned}. The command \texttt{python run\_nn.py --all --audit} runs every setting on both datasets and splits and computes the properties in Table~\ref{tab:splits}; \texttt{python run\_nn.py --summarize} writes the tables to \texttt{results/nearest\_neighbour/}.

\bibliography{references}
\bibliographystyle{apalike}

\end{document}